\documentclass[review]{elsarticle}

\usepackage{lipsum}
\makeatletter
\def\ps@pprintTitle{%
 \let\@oddhead\@empty
 \let\@evenhead\@empty
 \def\@oddfoot{}%
 \let\@evenfoot\@oddfoot}
\makeatother

\usepackage{hyperref,url}
\usepackage{rotating}
\usepackage{multirow}
\usepackage{amsmath}
\usepackage{xcolor}

\journal{arXiv}

\begin{document}
\sloppy

\begin{frontmatter}

\title{
Estimating brain age 
based on a healthy population 
with deep learning and structural MRI
} 

\author[1]{Xinyang Feng}
\author[2]{Zachary C. Lipton}
\author[1]{Jie Yang}
\author[3,4]{Scott A. Small\corref{cor1}}
\author[3,4]{Frank A. Provenzano\corref{cor1}}
\address[1]{Department of Biomedical Engineering, Columbia University}
\address[2]{Tepper School of Business, Carnegie Mellon University}
\address[3]{Department of Neurology, Columbia University}
\address[4]{Taub Institute for Research on Alzheimer's Disease and the Aging Brain, Columbia University}
\author{for the Alzheimer's Disease Neuroimaging Initiative\fnref{adni}}

\fntext[adni]{Data used in preparation of this article were partially obtained from the Alzheimer's Disease Neuroimaging Initiative (ADNI) database (adni.loni.usc.edu).
As such, the investigators within the ADNI contributed to the design and implementation of ADNI and/or provided data but did not participate in analysis or writing of this report.
A complete listing of ADNI investigators can be found at: \url{http://adni.loni.usc.edu/wp-content/uploads/how_to_apply/ADNI_Acknowledgement_List.pdf}}

\author{Australian Imaging Biomarkers and Lifestyle flagship study of ageing\fnref{aibl}}
\fntext[aibl]{Data used in the preparation of this article were partially obtained from the Australian Imaging Biomarkers and Lifestyle flagship study of ageing (AIBL)
funded by the Commonwealth Scientific and Industrial Research Organisation (CSIRO) which was made available at the ADNI database (www.loni.usc.edu/ADNI).
The AIBL researchers contributed data but did not participate in analysis or writing of this report. AIBL researchers are listed at www.aibl.csiro.au.}

\author{Frontotemporal Lobar Degeneration Neuroimaging Initiative\fnref{nifd}}
\fntext[nifd]{Data used in preparation of this article were partially obtained from the Frontotemporal Lobar Degeneration Neuroimaging Initiative (FTLDNI) database (http://4rtni-ftldni.ini.usc.edu).
The investigators at NIFD/FTLDNI contributed to the design and implementation of FTLDNI and/or provided data, but did not participate in analysis or writing of this report.}

\cortext[cor1]{Correspondence: Frank A. Provenzano (fap2005@cumc.columbia.edu) or Scott A. Small (sas68@cumc.columbia.edu)}

\begin{abstract}
Numerous studies have established that estimated brain age, as derived from statistical models trained on healthy populations, constitutes a valuable biomarker that is predictive of cognitive decline and various neurological diseases. In this work, we curate a large-scale heterogeneous dataset (N = 10,158, age range 18 - 97) of structural brain MRIs in a healthy population from multiple publicly-available sources, upon which we train a deep learning model for brain age estimation. The availability of the large-scale dataset enables a more uniform age distribution across adult life-span for effective age estimation with no bias toward certain age groups. We demonstrate that the age estimation accuracy, evaluated with mean absolute error (MAE) and correlation coefficient (\emph{r}), outperforms previously reported methods in both a hold-out test set reflective of the custom population (MAE = 4.06 years, \emph{r} = 0.970) and an independent life-span evaluation dataset (MAE = 4.21 years, \emph{r} = 0.960) on which a previous study has evaluated. We further demonstrate the utility of the estimated age in life-span aging analysis of cognitive functions. Furthermore, we conduct extensive ablation tests and employ feature-attribution techniques to analyze which regions contribute the most predictive value, demonstrating the prominence of the frontal lobe as well as pattern shift across life-span.

In summary, we achieve superior age estimation performance confirming the efficacy of deep learning and the added utility of training with data both in larger number and more uniformly distributed than in previous studies. We demonstrate the regional contribution to our brain age predictions through multiple routes and confirm the association of divergence between estimated and chronological brain age with neuropsychological measures. 

\end{abstract}


\end{frontmatter}


\section{Introduction}
\label{sec:introduction}

Age estimation is the task of estimating an individual's age
based on a set of other covariates.
In particular, a large body of research focuses 
on the task of predicting brain age based on imaging studies.
In addition to its utility in studying the aging process itself,
estimated age, as derived from models trained on a healthy population,
has emerged as a useful biomarker for diseases.
Specifically, the divergence of one's estimated age from chronological age
has been associated with various diseases, 
especially those thought to mimic an advanced age state.

Typically, models for estimating brain age are trained 
on datasets representing a normal aging population free of obvious disease.
When subsequently applied on (possibly abnormal) subjects,
predicted brain age has been linked to education 
and self-reported physical activity \cite{age_edu} 
and has been utilized to characterize diseases including
Alzheimer's Disease (AD) \cite{age_mci2ad}, schizophrenia \cite{age_schiz}, 
traumatic brain injury \cite{age_tbi}, etc., 
where deviation from normal aging trajectory occurs alongside disease state.

There are three components necessary to determine brain age with this paradigm: 
1) \textbf{covariates:} the measured brain characteristics which serve as inputs to the model, 
2) \textbf{dataset:} the precise cohort of normal aging subjects 
upon which the model is trained,
3) \textbf{model:} the precise algorithms
used to estimate the brain age given the available covariates.

While brain characteristics can be derived via many methods, 
neuroimaging is the most common and comprehensive way 
to characterize the ``brain state'' \emph{in vivo}.
Within neuroimaging, past studies have addressed
EEG (Electroencephalogram) \cite{age_eeg},
DTI (diffusion tensor imaging) \cite{age_dti},
and resting state BOLD fMRI (blood-oxygen-level dependent functional MRI) \cite{age_conn},
which reflect different physiological brain measures.
However, T1-weighted (T1w) structural MRI,
which reveals features of the underlying anatomical characteristics of the brain,
including gray and white matter delineation and gyral and sulcal patterns,
is the most common modality in brain age estimation research.
Imaging-derived neuroanatomical characteristics 
are biomarkers particularly sensitive to the aging process \cite{aging_consistency}.
Practically, as one most widely available and standardized neuroimaging modalities, 
T1w structural MRI can be easily acquired for a large population.
And within structural MRI domain, studies have used derived summary variables 
such as regional volumes or thickness estimation \cite{age_fs},
requiring additional software processing to generate such values, 
and also unprocessed MRI scans \cite{age_cnn}.

We believe that for these tools to be broadly useful,
they must be trained on adequately diverse datasets
that reflect the diversity of the populations 
on which the model might potentially be deployed.
In this study, we propose using a dataset aggregated 
from several publicly available multi-center neuroimaging datasets, 
representing a diverse healthy population.
This healthy study population is both larger in scale 
than those investigated by most previous age estimation studies 
and specifically designed to enable training age estimation 
models with an approximately uniform age distribution across the adult life span.

Lastly, given a study population and selected brain characteristics, 
the age is estimated based on statistical machine learning techniques.
Another way to formulate the age estimation problem 
is to extract generalizable features from the brain 
that best capture the chronological age of a person
provided the individual is undergoing a typical aging process 
that is present in a general healthy population.
Numerous traditional machine learning methods 
have been proposed for age estimation including
relevance vector machines \cite{age_diffeo,age_kernel},
Gaussian processes \cite{age_bayes,age_gp},
random forests \cite{age_rf},
hidden Markov models \cite{age_hmm},
and non-negative matrix factorization \cite{age_nmf}.

More modern deep learning based methods 
are especially well suited to this task 
provided enough training data,
and have been previously applied by \cite{age_cnn}
who demonstrated favorable performance. 
Interpretability is a critical aspect of deep neural network based method.
We seek not only to achieve predictive performance, 
but also to derive insights by understanding which features are most predictive.
In this work, we explore regional contributions in the regression task 
by considering both feature ablation experiments 
and an activation map based \emph{post hoc} interpretation method.

In summary, we utilize a 3D deep convolutional neural network based regression model
to estimate age using T1w structural MRI volumes
from a diverse multi-study population 
that is sampled with even age distribution across adult age span.
We achieved superior performance both in 
the hold-out test set from the same population 
and also in an independent life-span evaluation dataset.
The deviance of the estimated age from chronological age 
is linked to neuropsychological and neuromorphometric measures.
We further demonstrate the prominence of frontal lobe in brain age estimation, 
and the pattern variability across life-span.

\section{Method}
\label{sec:method}
In this section, we first describe the population 
and the experimental setup used in this study.
We then describe the MRI pre-processing steps
and the convolutional neural network used to estimate age.
Next, given our learned model for estimating brain age,
we present analysis that associates 
the divergence of the estimated age from the chronological brain age 
with neuropsychological and neuromorphometric measures.
Finally, we describe multiple modes of regionality analysis 
for identifying which regions are most predictive.
These consist of both ablation experiments 
and an extension of the class activation mapping method
for interpreting feature importance in deep networks 
to the regression setting.

\subsection{Study Population}
It is necessary to build an adequate neuroimaging dataset 
for age prediction in the full adult life span, 
especially given study recruitment criteria
the dataset is not necessarily evenly distributed across age.
Increased participation in open data consortia and imaging datasets 
greatly facilitates the collection of such data.
In this work, we collect more than 30,000 T1w MRI scans 
from multiple open neuroimaging datasets.
The list of the datasets used in this study 
with the full names and sources are listed 
in supplementary Table \ref{Table:dataset_name}.
Among those, we only include subjects 
with clear indication of normal neurological evaluations 
contingent on the individual data providers criteria 
for a subject that is considered free of disease. 
Specifically, we exclude subjects with any neurological or psychiatric disease, 
and also subjects with no available diagnosis label.
We also chose 18 as the minimum age to cover adulthood 
and also to avoid the neurodevelopmental stage.
This results in 10,158 MRI sessions from 6,142 unique subjects, 
the statistics are summarized in Table \ref{Table:dataset_stat}.

\newcommand{\sr}{\rule[0cm]{0pt}{0cm}}
\begin{sidewaystable}
	\centering
	\caption{\small{Dataset information}}\label{Table:dataset_stat}
	\vspace{1mm}
	\small{
		\renewcommand{\arraystretch}{1}
		\begin{tabular}{l|c|c|c|c|c|c}
			\hline
			\multirow{2}{*}{Dataset} & \multirow{2}{*}{$\mathrm{N_{sessions}}$} & \multirow{2}{*}{$\mathrm{N_{subjects}}$} & \multirow{2}{*}{age range} & \sr $\mathrm{age_{session}}$ & \sr $\mathrm{gender_{session}}$ & \sr $\mathrm{gender_{subject}}$\\
			& & & & \sr mean $\pm$ std & \sr M/F & \sr M/F \\ \hline \hline
			ADNI		&  2423	&  438	& 56.3 - 95.8 	& 76.64 $\pm$ \ 6.05 	& 1223/1192	& 217/221	\\ \hline 
			AIBL			&  781	&  457	& 60 - 92 		& 72.86 $\pm$ \ 6.53 	& 359/421	& 199/258  	\\ \hline
			NIFD 		&  428	&  136	& 36.9 - 85.2 	& 65.76 $\pm$ \ 7.66 	& 185/243	& 59/77 	\\ \hline
			IXI			&  561	&  561	& 20.0 - 86.3 	& 48.67 $\pm$ 16.47	& 248/313	& 248/313	\\ \hline
			BGSP		&  1566	&  1566	& 19 - 35 		& 21.54 $\pm$ \ 2.89 	& 661/905	& 661/905	\\ \hline
			Cam-CAN		&  652	&  652	& 18 - 88 		& 54.30 $\pm$ 18.59	& 322/330 	& 322/330	\\ \hline
			OASIS-1		&  316	&  316	& 18 - 94 		& 45.09 $\pm$ 23.90	& 119/197	& 119/197	\\ \hline
			OASIS-2		&  145	&  56	& 60 - 97 		& 76.54 $\pm$ \ 7.99 	& 52/93		& 19/37   	\\ \hline
			SALD		&  467	&  467	& 19 - 80 		& 45.07 $\pm$ 17.39	& 168/297	& 168/297	\\ \hline
			SLIM			&  972	&  561	& 18 - 28.5  	& 20.68 $\pm$ \ 1.40	& 425/547	& 244/317  	\\ \hline
			PPMI		&  130	&  74	& 30.6 - 81 	& 60.95 $\pm$ 10.85 & 91/39		& 48/26  	\\ \hline
			SchizConnect 	&  742	&  567	& 18 - 70  		& 35.36 $\pm$ 12.76 & 429/313	& 338/229  	\\ \hline
			DLBS		&  301	&  301	& 20.6 - 89.1 	& 53.62 $\pm$ 19.92 & 115/186	& 115/186	\\ \hline
			CoRR		&  1326	&  642	& 18 - 83  		& 28.78 $\pm$ 12.39 & 689/637	& 343/299  	\\ \hline \hline
			total			& 10158 & 6142	& 18 - 97		& 47.92 $\pm$ 24.89	& 4764/5783	& 2780/3362	\\ \hline
	\end{tabular}}
\end{sidewaystable}

\begin{figure}
	\centering
	\includegraphics[width=9cm]{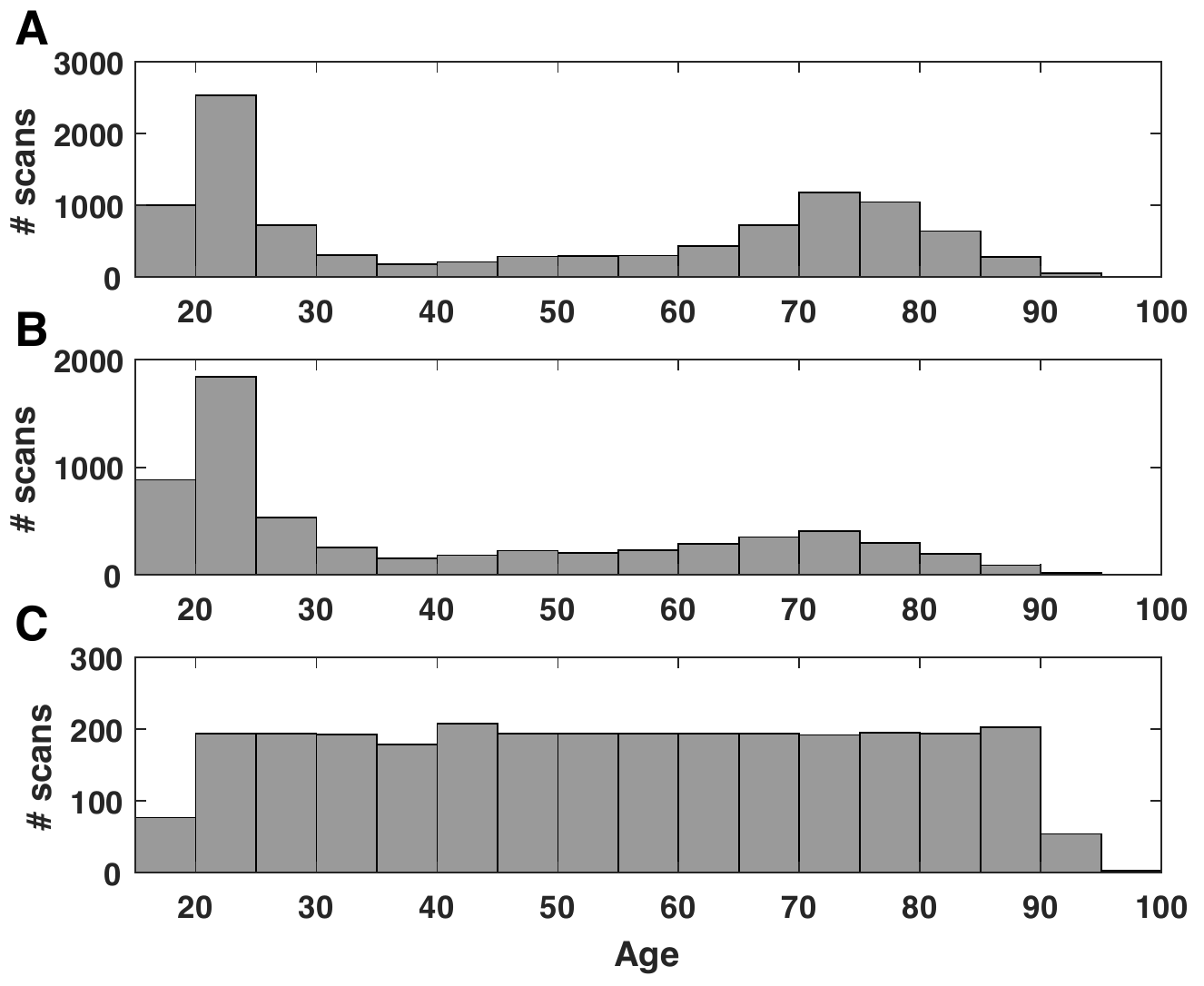}
	\label{fig:age}
	\caption{The age distribution of the study population. A) The age distribution of the raw dataset consisted of 10,158 MRI sessions;
	B) the age distribution of the dataset consisted of 6,142 unique subjects; C) the age distribution of the evenly sampled dataset.}
\end{figure}

However, as shown in Figure \ref{fig:age} A and B, 
where we illustrate the age distribution of 
the 10,158 sessions and 6,142 subjects respectively,
the age distribution of the population is poorly balanced.
Although there are studies covering the full age span 
including normal aging studies Cam-CAN 
\cite{camcan}, IXI, SALD \cite{sald}, DLBS \cite{dlbs},
OASIS-1 \cite{oasis_cs}, and  consortium-based studies 
such as CoRR \cite{corr}, SchizConnect \cite{schizconnnet},
many of the public imaging studies either focused 
on age-related disease in the elderly population 
including ADNI, AIBL \cite{aibl}, OASIS-2 \cite{oasis_l}, PPMI \cite{ppmi}, NIFD;
or on young subjects including BGSP \cite{bgsp}, SLIM \cite{slim}.
To alleviate the potential bias toward a certain age segment,
we sought to balance the age distribution 
that we ultimately use in the training population.

In this study, when constructing the dataset to be uniform across age-span, 
we adopt both `oversampling' and `undersampling'.
Briefly, we oversample 
subjects from 
age ranges with fewer subjects
by including the longitudinal follow-up sessions from the same subjects,
which could be regarded as a natural augmentation.
For age ranges with more subjects, we only include one scan per subject 
to increase the variability of the sample. 
If that number is still above the minimum across age bins,
we further undersample stratified on confounding factors 
including acquisition site and gender.

We stratify the populations into age bins and use the bin 
with the minimum number of subjects as the basis number.
The age bins used in this study are 
[18, 20), [20, 25), [25, 30), ... , [85, 90), [90, 100).

One interesting observation is that the [35, 40), [40, 45) 
are the two age bins with fewest number of subjects,
so we use the number of subjects in this age segment 
as the base level and allow repeated scans from same subjects.
The other age bins having multiple scans per subject 
are the two age bins at the tail end: [85, 90) and [90, 100], 
because of the relative lower number in these two age bins.
We undersample the subjects in all other bins.

The final dataset consists of 2,852 MRI sessions 
from 2,694 subjects covering age range 18-97, 
with the mean age 54.34 years old, standard deviation 21.16 years old.
The age distribution of the evenly-sampled adult age span dataset 
is shown in Figure \ref{fig:age} C. 

\subsection{Experimental setup}
We design training, validation and test sets of subjects 
for model training and evaluation.
We build a validation set 
reflecting of general MRI scan distribution and population distribution.
Similarly, a test set representative of the same population 
as the training and validation sets is important for evaluation.

Given this uniform age-distributed dataset we described in the section above,
we perform stratified split based on acquisition sites and gender 
\emph{within} each age bin: 8/10 as training set, 1/10 as validation set, 1/10 as test set,
ensuring non-overlapping subjects and similar distribution of age,
gender and acquisition sites in the training, validation and test sets.

We also evaluate our model by testing it on an independent test set: Cam-CAN.
Aiming to study the normal aging process through the adult life-span,
Cam-CAN provided even age distribution across adult life-span 
and was previously used as an independent testing sample 
in another age estimation study \cite{age_bayes}.
We note that the trained model should be applicable
to any similarly acquired structural neuroimaging within this age range.
However, the result on an independent sample, 
which usually reflects a homogeneous population or acquisition setting, 
might be over- or under-optimistic.

Additionally, we also perform a test-retest experiment using an independent dataset 
of three subjects scanned 40 times in 30 days \cite{maclaren} 
to evaluate the reproducibility of the model.

\subsection{MRI Preprocessing}
We apply basic pre-processing steps including nonparametric nonuniform intensity normalization (N3) based bias field correction \cite{sled1998n3}, 
brain extraction using FreeSurfer \cite{segonne2004skull},
and affine registration to the $1mm^3$ isotropic MNI152 brain template 
with trilinear interpolation using FSL FLIRT \cite{jenkinson2002flirt}. 
The dimension of the 3D volume is $182\times218\times182$ (LR$\times$AP$\times$SI). 

All the preprocessed scans were checked 
by a well-trained reviewer with visual inspection.
Scans having obvious and severe MRI artifacts, 
brain extraction failure or poor registration were excluded.

\subsection{Convolutional neural network}
We use a three-dimensional convolutional neural network (3D-CNN) 
regression model for age estimation,
with similar architecture as the 3D-CNN classification model for Alzheimer's disease classification used in \cite{addl}.
We follow a general CNN architecture 
similar to the VGG classification architecture \cite{vgg} 
with multiple interleaved convolutional blocks and max pooling layers 
and an increasing number of features along the depth. 
We replaced all 2D operations with 3D operations 
and included one fully-connected layer 
to reduce the number of parameters. 
We use convolutional layers with kernel size of $3\times3\times3$,
rectified linear unit (ReLU) as the activation function, 
and batch normalization (BN) layers before the activation functions. 
We flatten the output from the last convolutional layer
and feed into a fully-connected (FC) layer.
And the final activation is a linear operation 
instead of a softmax in classification tasks.
We include weight \textit{l2} regularization 
to prevent overfitting with a factor of $1.0$. 
The algorithm was optimized using Adam algorithm with
mean absolute error (MAE) loss function, and with a batch size of $5$.
The initial learning rate was tuned in the range from $1e-4$ to $1e-6$ 
including [$1$e-$4$, $5$e-$5$, $2$e-$5$, $1$e-$5$, $5$e-$6$, $2$e-$6$, $1$e-$6$] 
and was set at $2$e-$5$. 

\begin{figure}[!h]
	\centering
	\includegraphics[width=\linewidth]{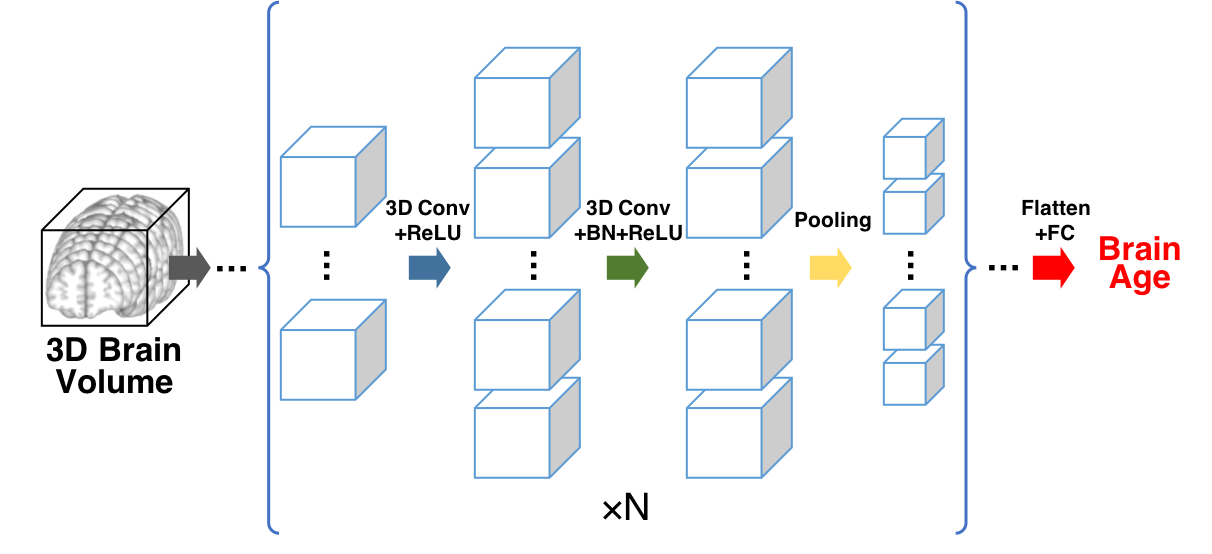}
	\caption{The convolution neural network architecture.
		The inputs are 3D brain volumes. 
		Each cube represents one 3D feature map.
		The size of the cube reflects the spatial dimension of the feature map, 
		the number of the cubes reflects the number of feature maps (channel dimension) at a specific depth. 
		The blue arrow denotes 3D convolution operation with rectified linear unit (ReLU), 
		the green arrow denotes 3D convolution followed by batch normalization (BN) and ReLU, 
		the yellow arrow denotes the max pooling operation. 
		The basic unit enclosed in the bracket is repeated $N=5$ times 
		with increasing number of features and decreasing spatial dimension. 
		The final convolutional output is flattened and fed into one fully-connected (FC) layer with linear output (red arrow), 
		generating the final brain age prediction.}
	\label{fig:cnn}
\end{figure}

We implemented the algorithm using Keras and TensorFlow. 
An illustration of the framework is shown in Figure \ref{fig:cnn}.
In this study, we use five (N in Figure \ref{fig:cnn}) stages. 
The feature dimension of the first layer is $16$ 
and increases by a factor of $2$ in each subsequent stage.
The optimal model is selected as the model with the lowest validation MAE.

\subsection{Comparison with model trained on unbalanced dataset}
We also trained the model using all scans from unique subjects 
(N = 6,142, Figure \ref{fig:age} B).
We applied the trained model on the independent Cam-CAN dataset 
and studied the distribution of MAE over chronological age groups.

Besides, a simple way to potentially correct the imbalance 
without adjusting the sampling of the dataset is re-weighting the samples,
specifically, we assign different weights to different sample, 
with the weights in proportional to the inverse 
of the frequency of specific age segments.
We note that while importance weighting is a principled statistical technique
with well-known effects in traditional statistical models,
its impact on the learned predictors in deep learning algorithms
are a subject of active inquiry \cite{byrd2019effect}.

\subsection{Neuropsychological and morphometric associations}
To test the utility of the estimated age in 
studying cognitive functions across adult life-span,
we evaluate the association between the summary scores 
of Benton face recognition test (BFRT) and the estimated age in Cam-CAN dataset.

Specifically, we use the signed difference of the estimated age and chronological age 
to reflect the deviance of individual brain age from their chronological age, 
and we refer to this value as $\mathrm{age_{diff}}$ hereinafter.
The BFRT is a commonly used neuropsychological instrument 
that can be easily and reliably administered in adult patients 
to test baseline visual memory and perception \cite{bfrt}.
We adopted the SubScore-1, SubScore-2, TotalScore (SubScore-1 + SubScore-2) 
as dependent variables in individual linear regression models incorporating gender, chronological age, $\mathrm{age_{diff}}$, and the interaction of chronological age and $\mathrm{age_{diff}}$:
\begin{equation*}
\mathrm{Score} \sim \beta_{\mathrm{age}} \mathrm{age} + \beta_{\mathrm{age_{diff}}} \mathrm{age_{diff}} + \beta_{\mathrm{age_{diff}\times age}} \mathrm{age \times age_{diff}} + \beta_{\mathrm{gender}} \mathrm{gender}
\end{equation*}

Additionally, we evaluate the association between ${age_{diff}}$ 
with cortical thickness generated using FreeSurfer \cite{fs_thickness}
by performing a partial correlation with gender and chronological age as covariates. 
FreeSurfer parcellates the cortex into 68 cortical regions.

\subsection{Age activation map}
Class activation mapping \cite{cam,gradcam} is a commonly-used method 
for interpreting the classification using CNNs 
and has been previously used in CNN based medical image analysis \cite{nam} 
to marry potential disease pathology with classification findings.
In this work, we use the idea of a class activation map in a regression setting 
by highlighting the small-valued gradient in grad-CAM framework. 
We use functions in the keras-vis package (\url{https://github.com/raghakot/keras-vis/}).
We generate the average activation map within each age group to investigate the age-specific pattern of underlying substrates for age estimation.

\subsection{Slice based age estimation}
Besides the \emph{post hoc} saliency map based activation map method,
we also propose an ablation analyses method focusing on part of the input data. 
We apply serial 2D CNNs for age estimation with the input 
being three consecutive slices along each axis.
By doing so, we take into consideration the inter-subject alignment precision 
(i.e. not using just one slice) 
and also the similarity among different channels (i.e. not using five slices).
The network architecture of the 2D CNN is the identical 
to the 3D CNN architecture described in the previous section 
with the 3D operations replaced with the corresponding 2D operations.
We report the predictive performance on various sets of 2D slices 
to analyze predictive importance.

\subsection{Lobe based age estimation}
Besides sliced based age estimation,
we propose using another more anatomically-informed way 
to study the regionality through ablation experiments at the lobar level.
The individual lobe masks were generated following a previous study \cite{addl}. 
The ages are estimated using each lobe individually.

\section{Results}
\label{sec:result}
\subsection{Age prediction}
In the hold-out test set, whose instances are distributed identically 
as the training and validation set, 
our model achieves an MAE of 4.06 years and correlation coefficient $r$ = 0.970.

\begin{figure}
	\centering
	\includegraphics[width=8cm]{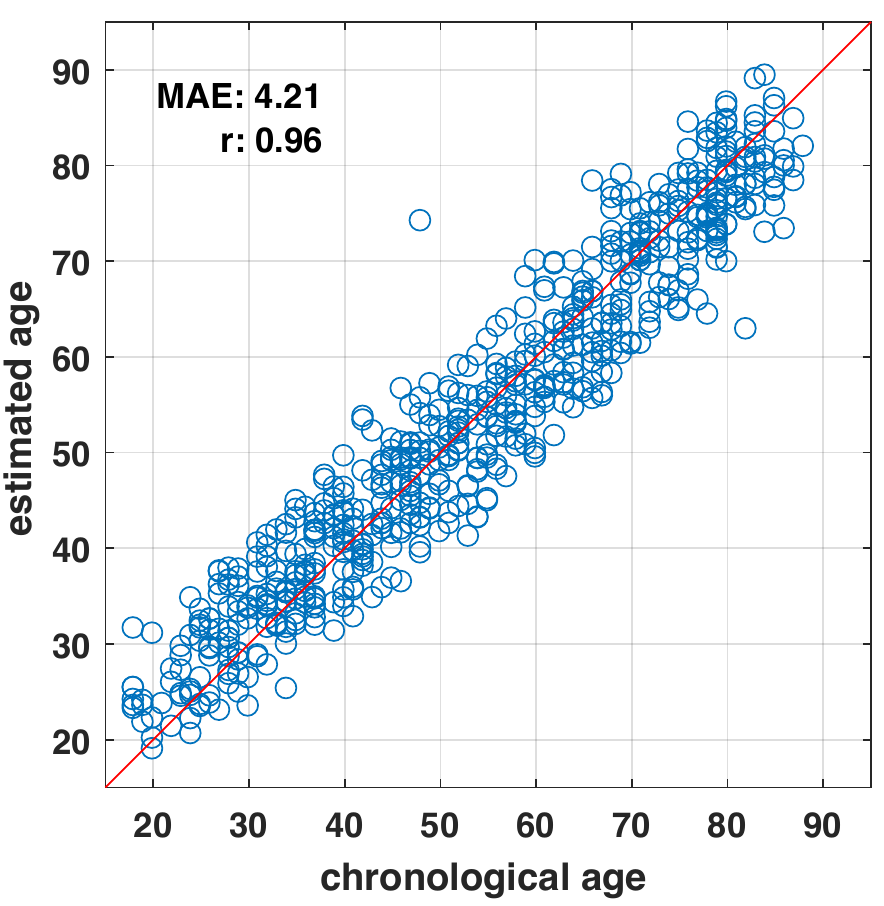}
	\label{fig:estage}
	\caption{The estimated age versus chronological age in an independent test set of adult life-span.}
\end{figure}

An independent normal aging life-span dataset---Cam-CAN---was 
tested in a previous brain age study \cite{age_bayes}.
In that study, when the Cam-CAN data were pre-processed 
with the optimal parameters selected from the independent training sample, 
their proposed model achieved an MAE of 6.08 years and correlation coefficient $r$ = 0.929.
We tested out our model in Cam-CAN study processing the T1w MRI images 
using the same pipeline as the other samples.
The relationship between the estimated age and chronological age in Cam-CAN 
is shown in Figure \ref{fig:estage}.
The correlation coefficient $r$ is 0.960, MAE is 4.21 years, 
which outperforms the result in the previous study \cite{age_bayes}.
We also observe two obvious ``outliers'' among the 652 subjects tested, 
further investigation is needed to pinpoint potential sources of error, 
either methodological, constitutional or true poor estimation.
The results demonstrate our proposed model 
achieves accurate estimation in all age segments.

\subsection{Reproducibility}
\begin{figure}
	\centering
	\includegraphics[width=10cm]{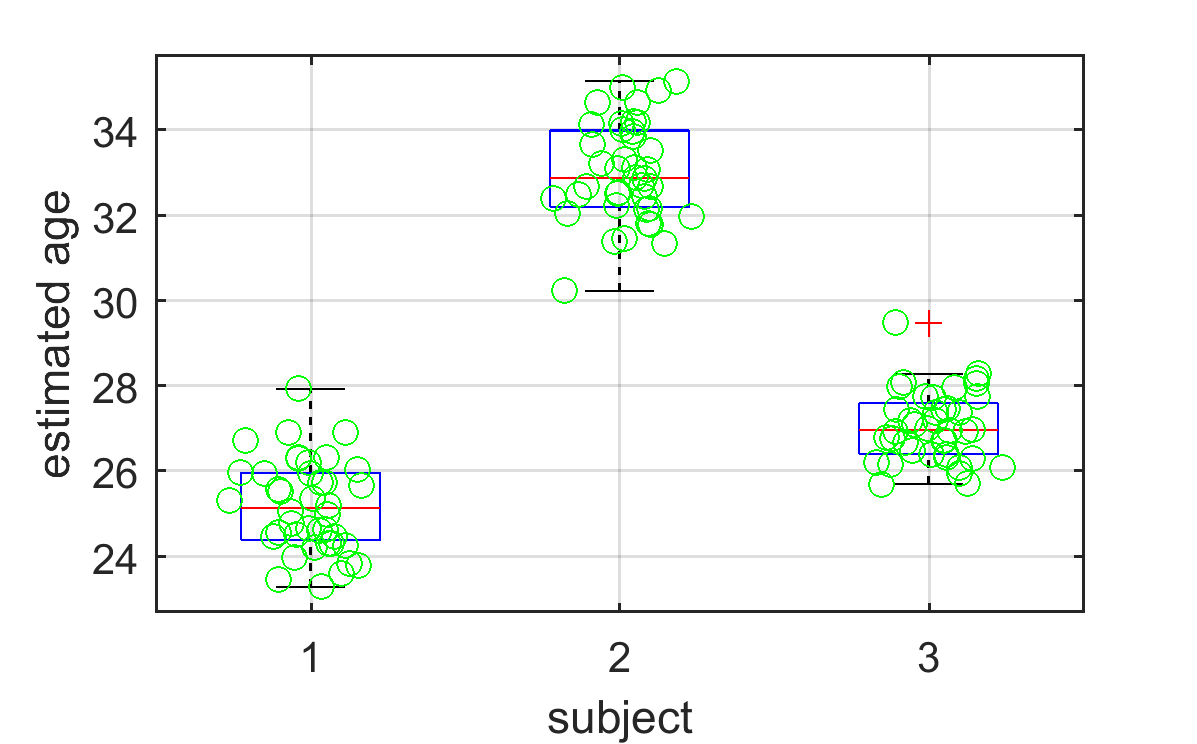}
	\caption{The distribution of predicted ages in test-retest scans.} \label{fig:repro_age}
\end{figure}

We evaluate the reproducibility of the algorithm in test-retest acquisitions.
We show the results in Table \ref{table:repro} and Figure \ref{fig:repro_age},
observing that there is a difference in the estimated age 
and actual reported chronological age that is consistent 
over the sessions with approximately a 1 year standard deviation.

\begin{table}
	\centering
	\caption{\small{Reproducibility experiment result}}\label{table:repro}
\begin{tabular}{l|c|c|c}
	\hline
	subject & actual age & predicted age mean & predicted age std \\ \hline
subj-1 & 26 & 25.19 & 1.07 \\ \hline
subj-2 & 31 & 33.02 & 1.14 \\ \hline
subj-3 & 30 & 27.06 & 0.81 \\ \hline				
\end{tabular}
\end{table}

\subsection{Comparison with results using nonuniform dataset}
\begin{figure}[!htbp]
	\centering
	\includegraphics[width=\linewidth]{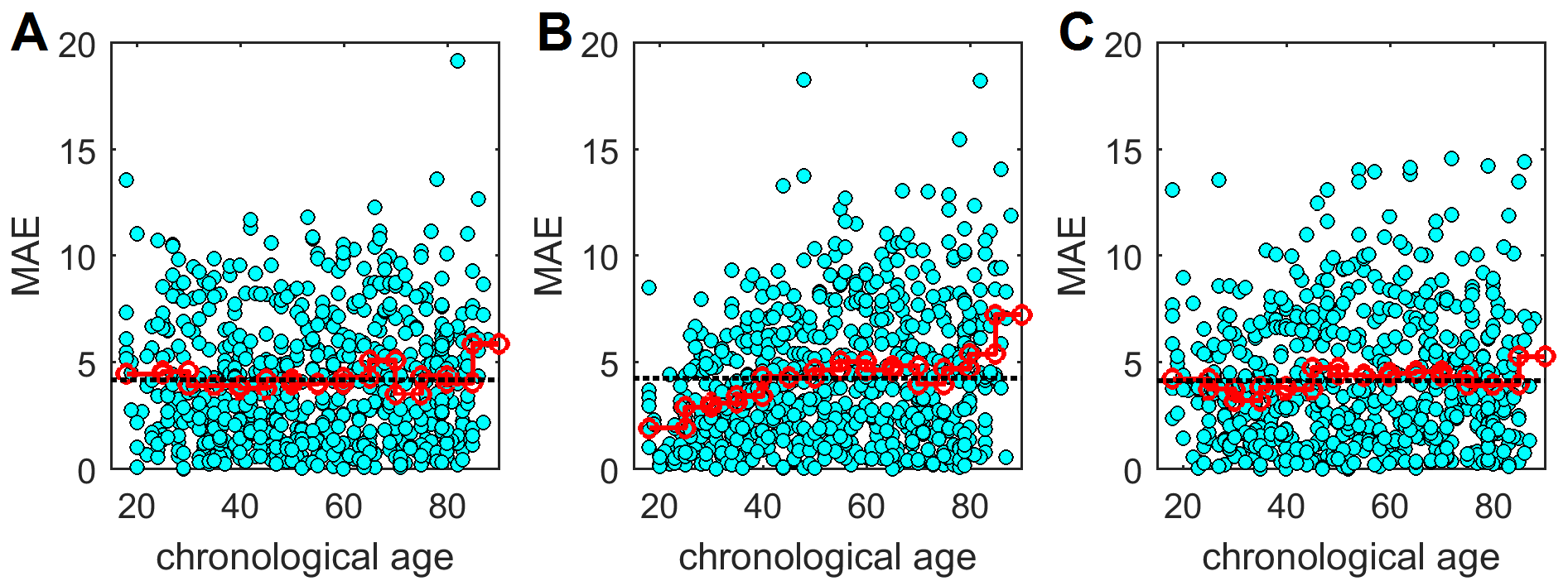}
	\caption{Distribution of MAE across life-span. (A) Age estimated using the balanced dataset.
	Each step in the red line indicate the MAE in that age group, the black dashed line indicates the overall MAE.
	(B) Age estimated using the nonuniform dataset. (C) Age estimated using the nonuniform dataset but with sample re-weighting. }
	\label{fig:maevsage}
\end{figure}

We compare the results using nonuniform dataset with the MAE performance in Cam-CAN dataset.
Using the nonuniform dataset achieves comparable overall MAE (4.27 years) as the balanced data. 
Re-weighting the samples helps slightly improves the MAE (4.17 years) 
than the balanced dataset despite using many more scans.
Additionally, we observe the MAE using the nonuniform dataset 
is not evenly distributed across life-span: 
MAE is lower in the young age with more abundant data, 
as shown in Figure \ref{fig:maevsage} (B).
This could introduce potential bias in life-span studies.
Using sample re-weighting (Figure \ref{fig:maevsage} (C)) alleviates the problem
and using balanced dataset generates generally even distribution across age-span (Figure \ref{fig:maevsage} (A)). 

\subsection{Neuropsychological and neuromorphometric association}
The association of the BFRT scores with the difference in the estimated age and the chronological age
and its interaction with chronological age are summarized in Table \ref{Table:benton}.
\begin{table}
	\centering
		\caption{Association with Benton face recognition scores}
		\label{Table:benton}
		\begin{tabular}{l|c|c|c|c|c|c}
		\hline
		\multirow{2}{*}{Score} & \multicolumn{2}{c}{age}  & \multicolumn{2}{|c|}{age diff}  & \multicolumn{2}{c}{age diff $\times$ age} \\ 
		\cline{2-7}
		& $\beta$ & $p$ & $\beta$ & $p$ & $\beta$ & $p$ \\
		\hline
		SubScore-1 & -2.51e-3 & 3.01e-6 & -0.0127 & 0.0297 & 2.62e-4 & 0.0102\\ \hline
		SubScore-2 & -0.0605 & 1.24e-32 & -0.157 & 2.90e-3 & 2.31e-3 & 0.0119\\ \hline
		TotalScore & -0.0630 & 5.07e-34 & -0.170 & 1.54e-3 & 2.57e-3 & 5.88e-3 \\ \hline
\end{tabular}
\end{table}

\begin{figure}[!htbp]
	\centering
	\includegraphics[width=11cm]{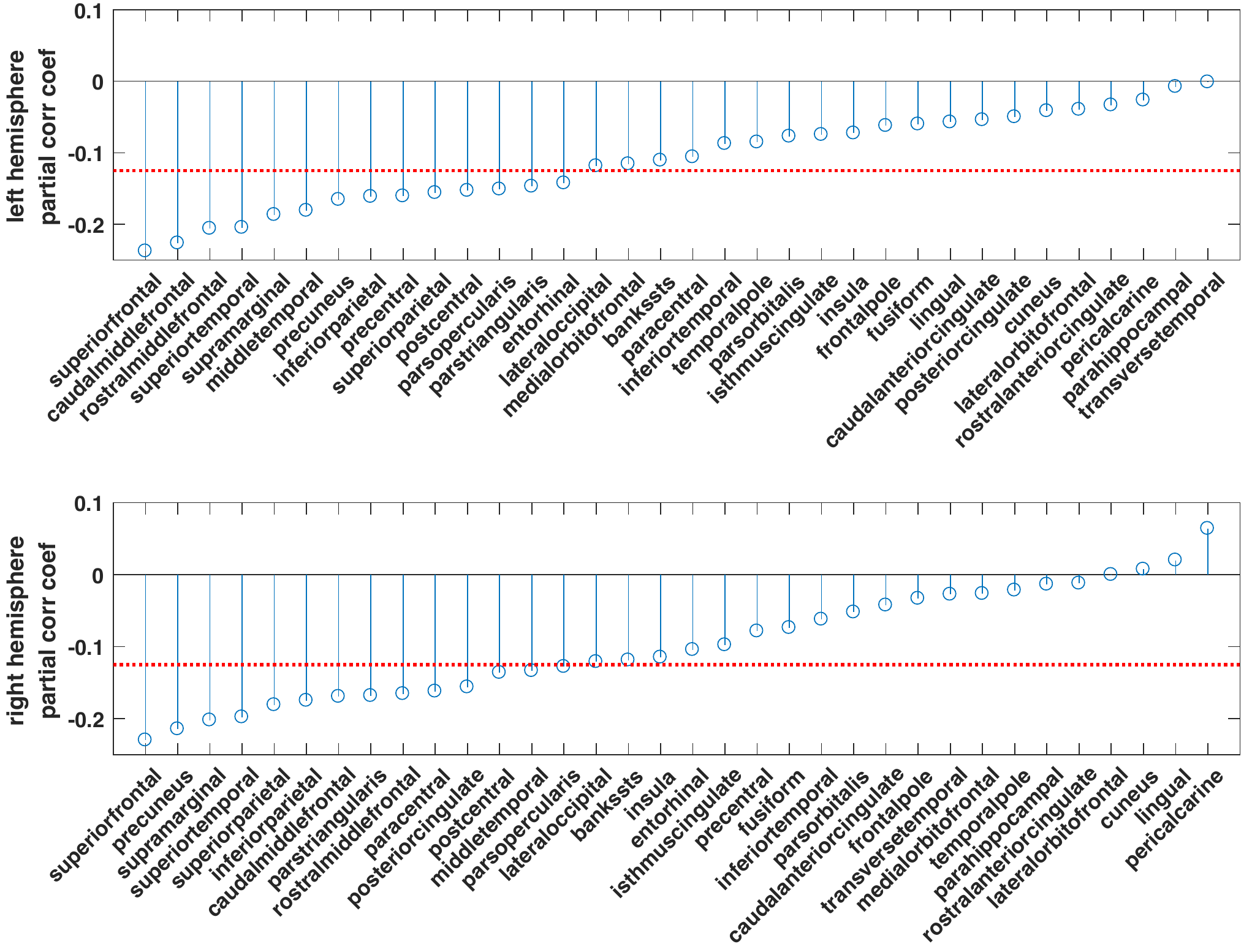}
	\caption{The partial correlation coefficients of $\mathrm{age_{diff}}$ and cortical thickness measures.
	The red dashed line indicates $\alpha=0.05$ under multiple comparison of N=34 regions.}
	\label{fig:thickness}
\end{figure}

The association of the cortical thickness measures with the $\mathrm{age_{diff}}$ is illustrated in Figure \ref{fig:thickness}.
The thickness of cortical regions are significantly associated with the $\mathrm{age_{diff}}$.
In addition, out of the 68 regions measured, 51 regions have a stronger correlation with the estimated age than the chronological age.
This is expected as the age estimated through structural MRI image is in principle more coupled to structural phenotypes.

\subsection{Age activation map}
\begin{figure}
	\centering
	\includegraphics[width=10cm]{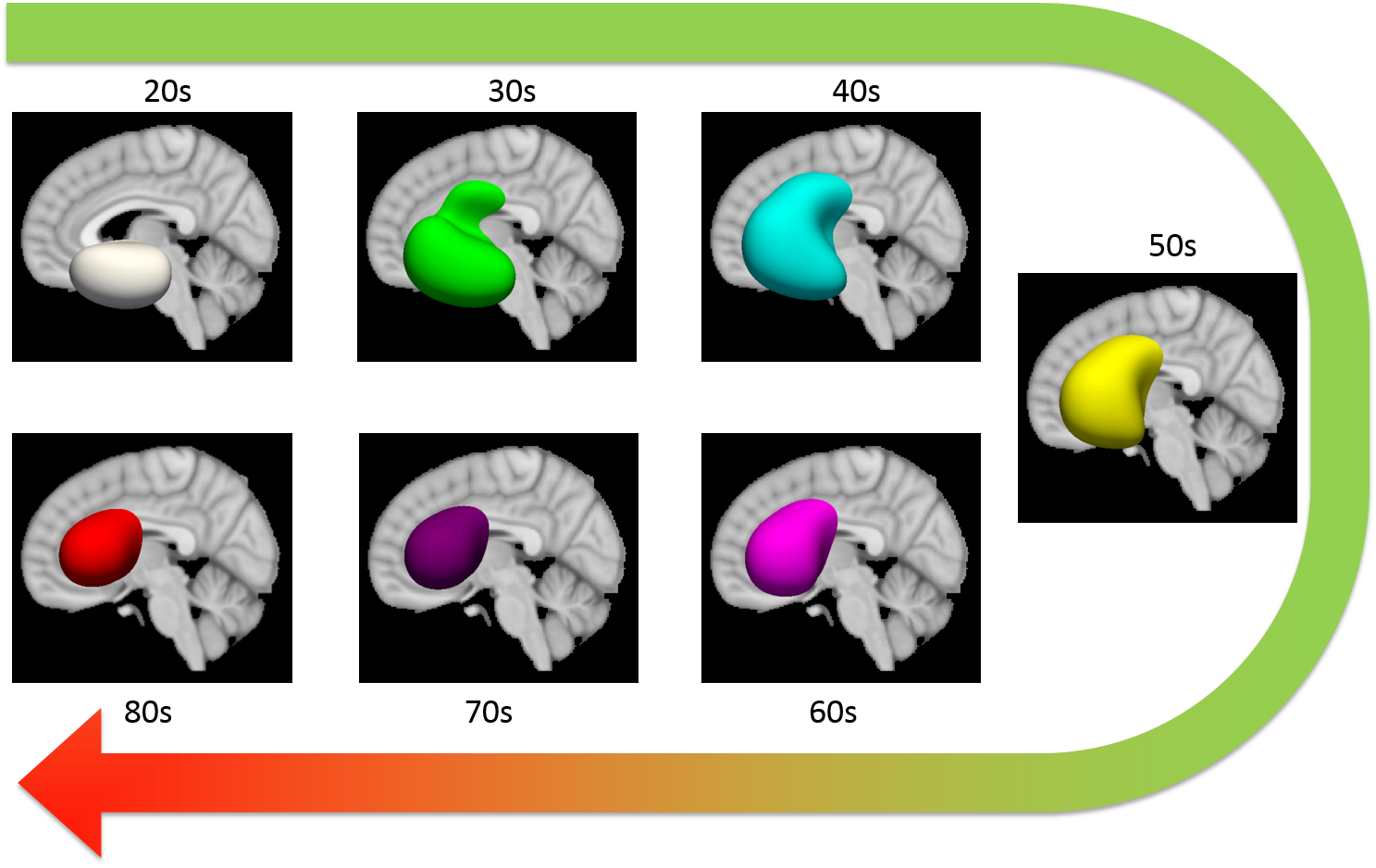}
	\caption{The 3D iso-surfaces (0.8) of the age activation maps at different age groups.} \label{fig:ram_3d}
\end{figure}

\begin{figure}
	\centering
	\includegraphics[width=5.988cm]{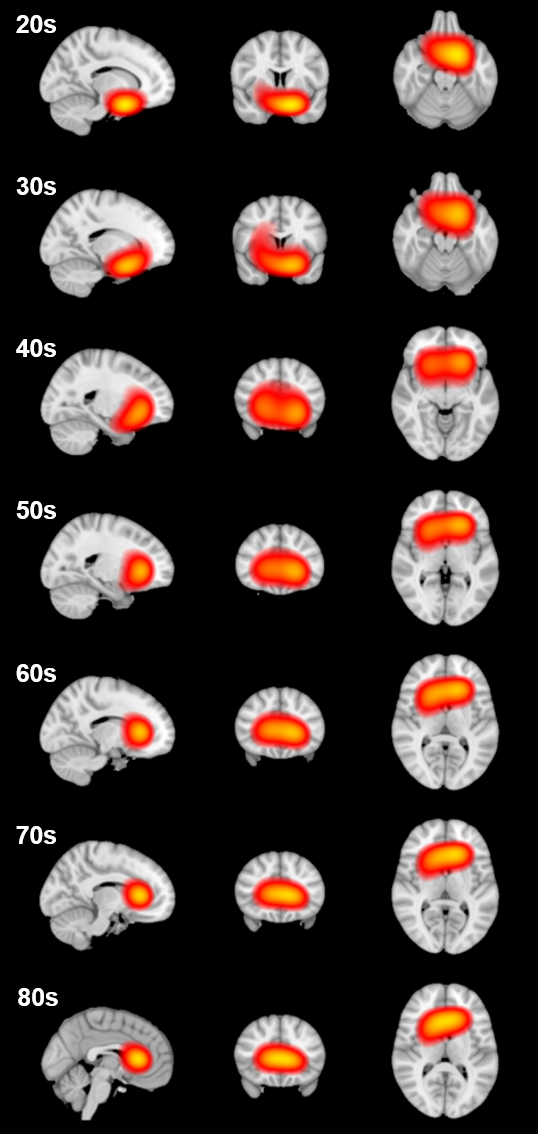}
	\includegraphics[width=6cm]{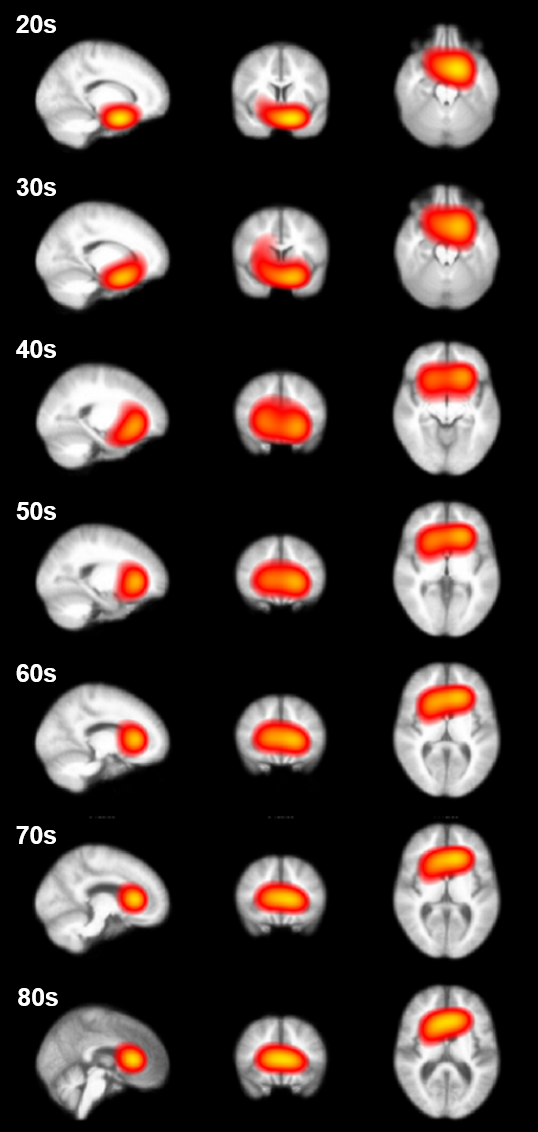}
	\caption{The age activation maps at different age groups overlaid on the
	(Left) MNI152 template, and
	(Right) average T1w image within each age group, both with threshold at 0.8.} \label{fig:ram_age_specific}
	\label{fig:ram}
\end{figure}

We expect the anatomical patterns characterizing different age groups to be different throughout lifetime but are consistent within a local age range.
Thus we generate and illustrate the age activation maps every 5 years, in the same way as preparing the dataset.
The 3D iso-surfaces of the average age activation maps are shown in Fig. \ref{fig:ram_3d}.
The average age activation maps overlaid on the MNI152 template are shown in Fig. \ref{fig:ram} (Left).
To accommodate the anatomical differences in different age groups,
we also generated average T1w image within each age group,
and overlaid the corresponding age activation map, as shown in Fig. \ref{fig:ram} (Right).

\subsection{Slice based age estimation}

\begin{figure}[!htbp]
	\centering
	\includegraphics[width=11cm]{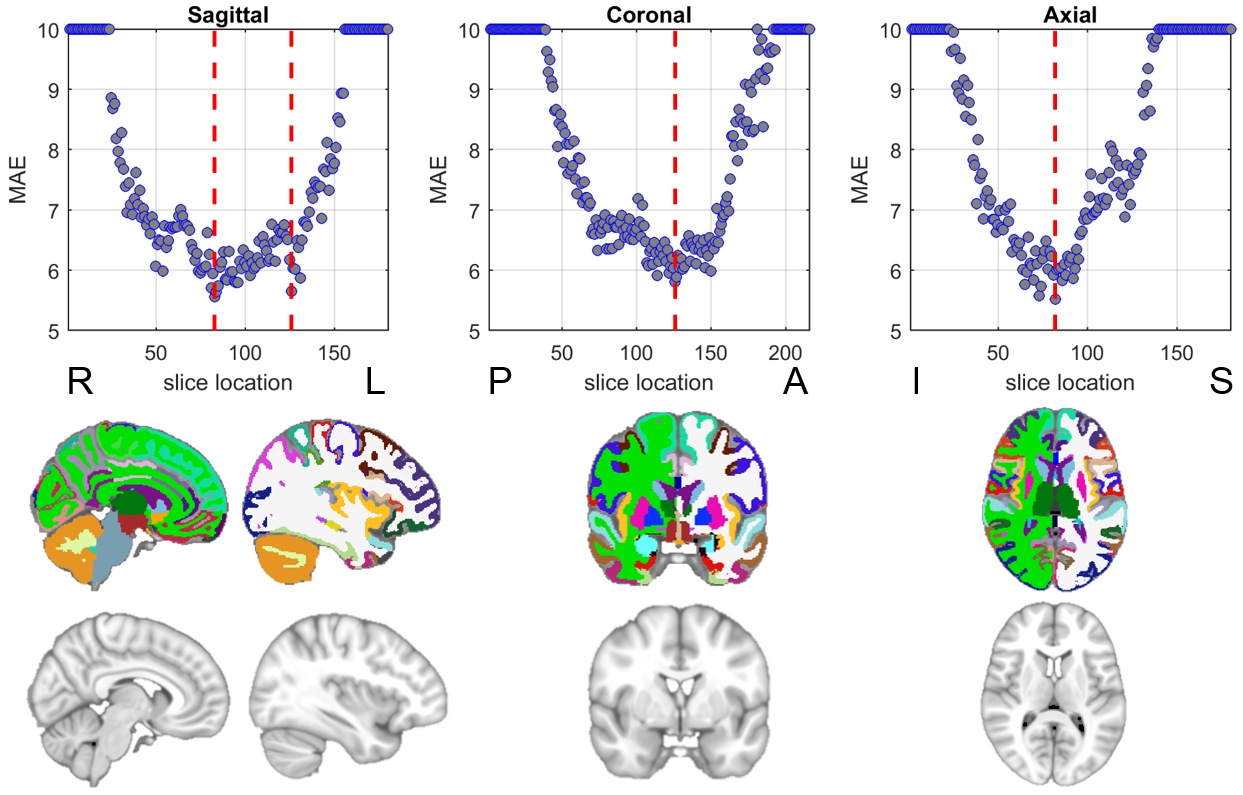}
	\caption{MRI 2D slice based age estimation. 
		(Top row) The mean absolute error (MAE) of the estimated age on the test set 
		using 2D slices at different locations, 
		the red lines indicate the location with lowest MAE. MAEs larger than 10 are capped to 10 for illustration purpose.
		(Bottom row) The illustration of slices at the red lines 
		in the top row from the MNI152 template and the corresponding segmentation
		(the colors follow the FreeSurfer color lookup table).}
	\label{fig:slice}
\end{figure}

The age estimation performance using 2D MRI slabs sliced at different coordinate planes is shown in Figure \ref{fig:slice}.
The slices with the best performance are also illustrated.

\subsection{Lobe based age estimation}
The MAE trained using different lobes are shown in Table \ref{Table:lobe},
where we show the best estimation performance achieved through frontal lobe.
And temporal lobe achieved marginally lower performance.
\begin{table}[!htbp]
	\centering
		\caption{Lobe based age estimation}
		\label{Table:lobe}
		\begin{tabular}{l|c|c|c|c|c|c}
		\hline
		Lobe	& Frontal	&	Temporal	& Parietal	& Occipital	& Cerebellum	& Whole-brain \\\hline
		MAE	&	5.33	&	5.81		&	6.37	& 7.66		& 6.2	0		& 4.06  \\\hline
\end{tabular}
\end{table}

\section{Discussion}
\label{sec:discussion}

In this study, a large heterogeneous dataset of structural neuroimaging 
across the adult lifespan was aggregated from multiple publicly available data sources.
A deep convolutional neural network was developed, 
trained and applied to predict the chronological age from the subject's scan.

This uniformly distributed dataset is able to achieve estimation 
without appreciable bias towards a certain age group,
while maintaining training efficiency compared with training on an nonuniform dataset.
While this study aims to examine the aging process across the adult life-span,
other studies aiming to examine the role of aging in other conditions 
would likely benefit from other specific training dataset inclusion criteria. 
For example, for studies of autism and prodromal psychosis 
where age of onset skews younger, studies would require the inclusion of subjects below 18,
and subjects in middle and old age would likely be less informative.

The regionality analysis in this study reveals patterns 
of neuroanatomical contributions of normal aging.
All analyses provide evidence for the prominence of frontal regions 
in all epochs of age estimation in the adult lifespan.
Frontal regions have been implicated in normal aging
through both neuropsychological studies and neuroimaging studies \cite{nn_frontal_aging, fdg_frontal_aging_chetelat, fdg_frontal_aging}.
In addition, the pattern shifts reflected in the age activation map based analysis
imply a frontal lobe-focused locus of age-related structural changes.
Neuropsychological evaluations targeting different cognitive domains and brain regions might help reveal the underlying complexity in the aging-process. 

Our analysis revealed the association between the divergence of estimated age from chronological age and BFRT performance.
This suggests the potential utility of the estimated age at normal aging evaluation,
in complement to other cognitive test and neuroimaging based measures. 
The utility relates to an open question of the aging process.
It still requires further validation how a scan's predicted deviance 
would reveal an individual's brain health status or even trigger clinical evaluations, 
since while inter-subject differences in normal aging process exist,
they are not well understood.
Moreover, it is unclear whether this inherent variability increases with age 
or stays constant 
throughout the life-span.

One consideration is that the set of subjects with no current pathological symptoms 
might have disease at asymptomatic prodromal state 
that have yet to manifest in the clinical evaluation, such as pre-clinical AD. 
Such work would require follow-up data to preclude disease for a given period of time 
such that they remain stable as controls. 
However, this was not available for all subjects,
and in a group of otherwise healthy individuals, 
it is likely that some healthy individuals do harbor occult disease.
Additionally, subjects were considered normal using criteria 
germane to the disease of focus in different studies. 

Given the diversity of the studies from which the model was trained, 
the protocols and acquisitions vary across both site and study. 
This in fact benefits this model compared to individual studies of unified aging 
because the variance of acquisition differences are included in the training, 
compared to training data where a more cohesive protocol is used. 
Such inclusion without appreciable loss of accuracy supports 
the use of this model on a wider variety of baseline scans.

\section{Conclusion}
\label{sec:conclusion}
This study has demonstrated the high accuracy of an age estimation framework
 using routine structural neuroimaging and a deep convolutional neural network,
 trained on a large-scale healthy population with uniform age distribution across adult age span.
While there is considerable interest and value in a highly accurate estimation model that can be applied to a wide variety of structural MRI protocols,
this study also demonstrates patterns of the regional contributions of aging across the lifespan using multiple methods,
agnostic of tissue classes, structure delineation or surface parcellation.
This framework can be used to generate meaningful quantitative and visual biomarkers for both aging studies and broad neuroimaging applications.
Further studies may examine the estimated age and the regional contributions
to see if they may be useful in understanding one's age divergence in the context of a particular disease or condition's pathology.

\section{Conflicts of interest}
\small{
FAP is a consultant for and has equity in Imij Technologies. SAS has equity in Imij technologies. XF, FAP and SAS have a patent application filed related to neuroimaging methods. 
}

\section{Acknowledgment}
\small
Data were provided in part by IXI, accessed from http://brain-development.org/ixi-dataset/.

Data were provided in part by OASIS. OASIS Cross-Sectional: Principal Investigators: D. Marcus, R, Buckner, J, Csernansky J. Morris; P50 AG05681, P01 AG03991, P01 AG026276, R01 AG021910, P20 MH071616, U24 RR021382
OASIS: Longitudinal: Principal Investigators: D. Marcus, R, Buckner, J. Csernansky, J. Morris; P50 AG05681, P01 AG03991, P01 AG026276, R01 AG021910, P20 MH071616, U24 RR021382.

Data were provided in part by the Brain Genomics Superstruct Project of Harvard University and the Massachusetts General Hospital, (Principal Investigators: Randy Buckner, Joshua Roffman, and Jordan Smoller), with support from the Center for Brain Science Neuroinformatics Research Group, the Athinoula A. Martinos Center for Biomedical Imaging, and the Center for Human Genetic Research. 20 individual investigators at Harvard and MGH generously contributed data to the overall project.

Data used in preparation of this article were obtained from the Alzheimer's Disease Neuroimaging Initiative (ADNI) database (adni.loni.usc.edu). As such, the investigators within the ADNI contributed to the design and implementation of ADNI and/or provided data but did not participate in analysis or writing of this report. A complete listing of ADNI investigators can be found at: \url{http://adni.loni.usc.edu/wp-content/uploads/how_to_apply/ADNI_Acknowledgement_List.pdf}

Data collection and sharing for this project was funded by the Alzheimer's Disease Neuroimaging Initiative (ADNI) (National Institutes of Health Grant U01 AG024904) and DOD ADNI (Department of Defense award number W81XWH-12-2-0012). ADNI is funded by the National Institute on Aging, the National Institute of Biomedical Imaging and Bioengineering, and through generous contributions from the following: AbbVie, Alzheimer’s Association; Alzheimer’s Drug Discovery Foundation; Araclon Biotech; BioClinica, Inc.; Biogen; Bristol-Myers Squibb Company; CereSpir, Inc.; Cogstate; Eisai Inc.; Elan Pharmaceuticals, Inc.; Eli Lilly and Company; EuroImmun; F. Hoffmann-La Roche Ltd and its affiliated company Genentech, Inc.; Fujirebio; GE Healthcare; IXICO Ltd.; Janssen Alzheimer Immunotherapy Research \& Development, LLC.; Johnson \& Johnson Pharmaceutical Research \& Development LLC.; Lumosity; Lundbeck; Merck \& Co., Inc.; Meso Scale Diagnostics, LLC.; NeuroRx Research; Neurotrack Technologies; Novartis Pharmaceuticals Corporation; Pfizer Inc.; Piramal Imaging; Servier; Takeda Pharmaceutical Company; and Transition Therapeutics. The Canadian Institutes of Health Research is providing funds to support ADNI clinical sites in Canada. Private sector contributions are facilitated by the Foundation for the National Institutes of Health (www.fnih.org). The grantee organization is the Northern California Institute for Research and Education, and the study is coordinated by the Alzheimer’s Therapeutic Research Institute at the University of Southern California. ADNI data are disseminated by the Laboratory for Neuro Imaging at the University of Southern California.

Data used in the preparation of this article was obtained from the Australian Imaging Biomarkers and Lifestyle flagship study of ageing (AIBL) funded by the Commonwealth Scientific and Industrial Research Organisation (CSIRO) which was made available at the ADNI database (www.loni.usc.edu/ADNI). The AIBL researchers contributed data but did not participate in analysis or writing of this report. AIBL researchers are listed at www.aibl.csiro.au.

Data used in preparation of this article were obtained from the Frontotemporal Lobar Degeneration Neuroimaging Initiative (FTLDNI) database (http://4rtni-ftldni.ini.usc.edu). The investigators at NIFD/FTLDNI contributed to the design and implementation of FTLDNI and/or provided data, but did not participate in analysis or writing of this report (unless otherwise listed).

Data collection and sharing for this project was funded by the Frontotemporal Lobar Degeneration Neuroimaging Initiative (National Institutes of Health Grant R01 AG032306). The study is coordinated through the University of California, San Francisco, Memory and Aging Center. FTLDNI data are disseminated by the Laboratory for Neuro Imaging at the University of Southern California.

Data used in the preparation of this article were obtained from the Parkinson's Progression Markers Initiative (PPMI) database (www.ppmi-info.org/data). For up-to-date information on the study, visit www.ppmi-info.org.
PPMI - a public-private partnership - is funded by the Michael J. Fox Foundation for Parkinson's Research and funding partners, the full names of all of the PPMI funding partners can found at www.ppmi-info.org/fundingpartners.

Data collection and sharing for this project was provided by the Cambridge Centre for Ageing and Neuroscience (Cam-CAN). CamCAN funding was provided by the UK Biotechnology and Biological Sciences Research Council (grant number BB/H008217/1), together with support from the UK Medical Research Council and University of Cambridge, UK.

Data used in preparation of this article were obtained from the SchizConnect database (http://schizconnect.org) As such, the investigators within SchizConnect contributed to the design and implementation of SchizConnect and/or provided data but did not participate in analysis or writing of this report.

Data collection and sharing for this project was funded by NIMH cooperative agreement 1U01MH097435.



\bibliography{age_estimation}

\begin{thebibliography}{10}
\expandafter\ifx\csname url\endcsname\relax
  \def\url#1{\texttt{#1}}\fi
\expandafter\ifx\csname urlprefix\endcsname\relax\def\urlprefix{URL }\fi
\expandafter\ifx\csname href\endcsname\relax
  \def\href#1#2{#2} \def\path#1{#1}\fi

\bibitem{age_edu}
J.~Steffener, C.~Habeck, D.~O'Shea, Q.~Razlighi, L.~Bherer, Y.~Stern,
  Differences between chronological and brain age are related to education and
  self-reported physical activity, Neurobiology of Aging 40 (2016) 138--144.
\newblock \href
  {http://dx.doi.org/http://dx.doi.org/10.1016/j.neurobiolaging.2016.01.014}
  {\path{doi:http://dx.doi.org/10.1016/j.neurobiolaging.2016.01.014}}.

\bibitem{age_mci2ad}
C.~Gaser, K.~Franke, S.~Klöppel, N.~Koutsouleris, H.~Sauer, I.~Alzheimer's
  Disease~Neuroimaging, {BrainAGE} in mild cognitive impaired patients:
  Predicting the conversion to {A}lzheimer's disease, PLOS ONE 8~(6) (2013)
  e67346.
\newblock \href
  {http://dx.doi.org/https://doi.org/10.1371/journal.pone.0067346}
  {\path{doi:https://doi.org/10.1371/journal.pone.0067346}}.

\bibitem{age_schiz}
H.~G. Schnack, N.~E. van Haren, M.~Nieuwenhuis, H.~E.~H. Pol, W.~Cahn, R.~S.
  Kahn, Accelerated brain aging in schizophrenia: A longitudinal pattern
  recognition study, American Journal of Psychiatry 173~(6) (2016) 607--616.
\newblock \href
  {http://dx.doi.org/https://doi.org/10.1176/appi.ajp.2015.15070922}
  {\path{doi:https://doi.org/10.1176/appi.ajp.2015.15070922}}.

\bibitem{age_tbi}
J.~H. Cole, R.~Leech, D.~J. Sharp, I.~for~the Alzheimer's Disease~Neuroimaging,
  Prediction of brain age suggests accelerated atrophy after traumatic brain
  injury, Annals of Neurology 77~(4) (2015) 571--581.
\newblock \href {http://dx.doi.org/https://doi.org/10.1002/ana.24367}
  {\path{doi:https://doi.org/10.1002/ana.24367}}.

\bibitem{age_eeg}
O.~Al~Zoubi, C.~Ki~Wong, R.~T. Kuplicki, H.-w. Yeh, A.~Mayeli, H.~Refai,
  M.~Paulus, J.~Bodurka, Predicting age from brain {EEG} signals—a machine
  learning approach, Frontiers in Aging Neuroscience 10~(184).
\newblock \href {http://dx.doi.org/http://dx.doi.org/10.3389/fnagi.2018.00184}
  {\path{doi:http://dx.doi.org/10.3389/fnagi.2018.00184}}.

\bibitem{age_dti}
B.~Mwangi, K.~M. Hasan, J.~C. Soares, Prediction of individual subject's age
  across the human lifespan using diffusion tensor imaging: A machine learning
  approach, NeuroImage 75~(0) (2013) 58--67.
\newblock \href
  {http://dx.doi.org/https://doi.org/10.1016/j.neuroimage.2013.02.055}
  {\path{doi:https://doi.org/10.1016/j.neuroimage.2013.02.055}}.

\bibitem{age_conn}
H.~Li, T.~D. Satterthwaite, Y.~Fan, Brain age prediction based on resting-state
  functional connectivity patterns using convolutional neural networks, in:
  International Symposium on Biomedical Imaging (ISBI), 2018, pp. 101--104.
\newblock \href {http://dx.doi.org/https://doi.org/10.1109/ISBI.2018.8363532}
  {\path{doi:https://doi.org/10.1109/ISBI.2018.8363532}}.

\bibitem{aging_consistency}
K.~B. Walhovd, L.~T. Westlye, I.~Amlien, T.~Espeseth, I.~Reinvang, N.~Raz,
  I.~Agartz, D.~H. Salat, D.~N. Greve, B.~Fischl, A.~M. Dale, A.~M. Fjell,
  Consistent neuroanatomical age-related volume differences across multiple
  samples, Neurobiology of Aging 32~(5) (2011) 916--932.
\newblock \href
  {http://dx.doi.org/https://doi.org/10.1016/j.neurobiolaging.2009.05.013}
  {\path{doi:https://doi.org/10.1016/j.neurobiolaging.2009.05.013}}.

\bibitem{age_fs}
S.~A. Valizadeh, J.~Hänggi, S.~Mérillat, L.~Jäncke, Age prediction on the
  basis of brain anatomical measures, Human Brain Mapping (2016) n/a--n/a\href
  {http://dx.doi.org/http://dx.doi.org/10.1002/hbm.23434}
  {\path{doi:http://dx.doi.org/10.1002/hbm.23434}}.

\bibitem{age_cnn}
J.~H. Cole, R.~P.~K. Poudel, D.~Tsagkrasoulis, M.~W.~A. Caan, C.~Steves, T.~D.
  Spector, G.~Montana, Predicting brain age with deep learning from raw imaging
  data results in a reliable and heritable biomarker, NeuroImage\href
  {http://dx.doi.org/https://doi.org/10.1016/j.neuroimage.2017.07.059}
  {\path{doi:https://doi.org/10.1016/j.neuroimage.2017.07.059}}.

\bibitem{age_diffeo}
J.~Ashburner, A fast diffeomorphic image registration algorithm, NeuroImage
  38~(1) (2007) 95--113.
\newblock \href
  {http://dx.doi.org/https://doi.org/10.1016/j.neuroimage.2007.07.007}
  {\path{doi:https://doi.org/10.1016/j.neuroimage.2007.07.007}}.

\bibitem{age_kernel}
K.~Franke, G.~Ziegler, S.~Klöppel, C.~Gaser, Estimating the age of healthy
  subjects from {T1-weighted MRI} scans using kernel methods: Exploring the
  influence of various parameters, NeuroImage 50~(3) (2010) 883--892.
\newblock \href
  {http://dx.doi.org/https://doi.org/10.1016/j.neuroimage.2010.01.005}
  {\path{doi:https://doi.org/10.1016/j.neuroimage.2010.01.005}}.

\bibitem{age_bayes}
J.~Lancaster, R.~Lorenz, R.~Leech, J.~H. Cole, Bayesian optimization for
  neuroimaging pre-processing in brain age classification and prediction,
  Frontiers in Aging Neuroscience 10~(28).
\newblock \href {http://dx.doi.org/http://dx.doi.org/10.3389/fnagi.2018.00028}
  {\path{doi:http://dx.doi.org/10.3389/fnagi.2018.00028}}.

\bibitem{age_gp}
B.~Gutierrez~Becker, T.~Klein, C.~Wachinger, Gaussian process uncertainty in
  age estimation as a measure of brain abnormality, NeuroImage 175 (2018)
  246--258.
\newblock \href
  {http://dx.doi.org/https://doi.org/10.1016/j.neuroimage.2018.03.075}
  {\path{doi:https://doi.org/10.1016/j.neuroimage.2018.03.075}}.

\bibitem{age_rf}
E.~Konukoglu, B.~Glocker, D.~Zikic, A.~Criminisi, Neighbourhood approximation
  using randomized forests, Medical Image Analysis 17~(7) (2013) 790--804.
\newblock \href {http://dx.doi.org/https://doi.org/10.1016/j.media.2013.04.013}
  {\path{doi:https://doi.org/10.1016/j.media.2013.04.013}}.

\bibitem{age_hmm}
B.~Wang, T.~D. Pham, Mri-based age prediction using hidden {Markov} models,
  Journal of Neuroscience Methods 199~(1) (2011) 140--145.
\newblock \href
  {http://dx.doi.org/https://doi.org/10.1016/j.jneumeth.2011.04.022}
  {\path{doi:https://doi.org/10.1016/j.jneumeth.2011.04.022}}.

\bibitem{age_nmf}
D.~P. Varikuti, S.~Genon, A.~Sotiras, H.~Schwender, F.~Hoffstaedter, K.~R.
  Patil, C.~Jockwitz, S.~Caspers, S.~Moebus, K.~Amunts, C.~Davatzikos, S.~B.
  Eickhoff, Evaluation of non-negative matrix factorization of grey matter in
  age prediction, NeuroImage 173 (2018) 394--410.
\newblock \href
  {http://dx.doi.org/https://doi.org/10.1016/j.neuroimage.2018.03.007}
  {\path{doi:https://doi.org/10.1016/j.neuroimage.2018.03.007}}.

\bibitem{camcan}
J.~R. Taylor, N.~Williams, R.~Cusack, T.~Auer, M.~A. Shafto, M.~Dixon, L.~K.
  Tyler, C.~A.~N. Cam, R.~N. Henson, The {Cambridge Centre for Ageing and
  Neuroscience (Cam-CAN)} data repository: Structural and functional {MRI},
  {MEG}, and cognitive data from a cross-sectional adult lifespan sample,
  NeuroImage 144 (2017) 262--269.
\newblock \href
  {http://dx.doi.org/https://doi.org/10.1016/j.neuroimage.2015.09.018}
  {\path{doi:https://doi.org/10.1016/j.neuroimage.2015.09.018}}.

\bibitem{sald}
D.~Wei, K.~Zhuang, Q.~Chen, W.~Yang, W.~Liu, K.~Wang, J.~Sun, J.~Qiu,
  Structural and functional {MRI} from a cross-sectional southwest university
  adult lifespan dataset ({SALD}), bioRxiv (2018) 177279\href
  {http://dx.doi.org/https://doi.org/10.1101/177279}
  {\path{doi:https://doi.org/10.1101/177279}}.

\bibitem{dlbs}
K.~Rodrigue, K.~Kennedy, M.~Devous, J.~Rieck, A.~Hebrank, R.~Diaz-Arrastia,
  D.~Mathews, D.~Park, β-amyloid burden in healthy aging: regional
  distribution and cognitive consequences, Neurology 78~(6) (2012) 387--395.
\newblock \href
  {http://dx.doi.org/https://doi.org/10.1212/WNL.0b013e318245d295}
  {\path{doi:https://doi.org/10.1212/WNL.0b013e318245d295}}.

\bibitem{oasis_cs}
D.~S. Marcus, T.~H. Wang, J.~Parker, J.~G. Csernansky, J.~C. Morris, R.~L.
  Buckner, {Open Access Series of Imaging Studies (OASIS)}: cross-sectional
  {MRI} data in young, middle aged, nondemented, and demented older adults,
  Journal of cognitive neuroscience 19~(9) (2007) 1498--1507.
\newblock \href {http://dx.doi.org/https://doi.org/10.1162/jocn.2007.19.9.1498}
  {\path{doi:https://doi.org/10.1162/jocn.2007.19.9.1498}}.

\bibitem{corr}
X.-N. Zuo, J.~S. Anderson, P.~Bellec, R.~M. Birn, B.~B. Biswal, J.~Blautzik,
  J.~C. Breitner, R.~L. Buckner, V.~D. Calhoun, F.~X. Castellanos, An open
  science resource for establishing reliability and reproducibility in
  functional connectomics, Scientific data 1 (2014) 140049.
\newblock \href {http://dx.doi.org/https://doi.org/10.1038/sdata.2014.49}
  {\path{doi:https://doi.org/10.1038/sdata.2014.49}}.

\bibitem{schizconnnet}
L.~Wang, K.~I. Alpert, V.~D. Calhoun, D.~J. Cobia, D.~B. Keator, M.~D. King,
  A.~Kogan, D.~Landis, M.~Tallis, M.~D. Turner, S.~G. Potkin, J.~A. Turner,
  J.~L. Ambite, {SchizConnect}: Mediating neuroimaging databases on
  schizophrenia and related disorders for large-scale integration, NeuroImage
  124~(Pt B) (2016) 1155--67.
\newblock \href
  {http://dx.doi.org/https://doi.org/10.1016/j.neuroimage.2015.06.065}
  {\path{doi:https://doi.org/10.1016/j.neuroimage.2015.06.065}}.

\bibitem{aibl}
K.~A. Ellis, A.~I. Bush, D.~Darby, D.~De~Fazio, J.~Foster, P.~Hudson, N.~T.
  Lautenschlager, N.~Lenzo, R.~N. Martins, P.~Maruff, C.~Masters, A.~Milner,
  K.~Pike, C.~Rowe, G.~Savage, C.~Szoeke, K.~Taddei, V.~Villemagne,
  M.~Woodward, D.~Ames, The {Australian Imaging, Biomarkers and Lifestyle
  (AIBL)} study of aging: methodology and baseline characteristics of 1112
  individuals recruited for a longitudinal study of {A}lzheimer's disease,
  International Psychogeriatrics 21~(4) (2009) 672--687.
\newblock \href {http://dx.doi.org/https://doi.org/10.1017/S1041610209009405}
  {\path{doi:https://doi.org/10.1017/S1041610209009405}}.

\bibitem{oasis_l}
D.~S. Marcus, A.~F. Fotenos, J.~G. Csernansky, J.~C. Morris, R.~L. Buckner,
  Open access series of imaging studies: longitudinal {MRI} data in nondemented
  and demented older adults, Journal of cognitive neuroscience 22~(12) (2010)
  2677--2684.
\newblock \href {http://dx.doi.org/https://doi.org/10.1162/jocn.2009.21407}
  {\path{doi:https://doi.org/10.1162/jocn.2009.21407}}.

\bibitem{ppmi}
K.~Marek, D.~Jennings, S.~Lasch, A.~Siderowf, C.~Tanner, T.~Simuni, C.~Coffey,
  K.~Kieburtz, E.~Flagg, S.~Chowdhury, et~al., The {Parkinson Progression
  Marker Initiative (PPMI)}, Progress in Neurobiology 95~(4) (2011) 629--635.
\newblock \href
  {http://dx.doi.org/http://dx.doi.org/10.1016/j.pneurobio.2011.09.005}
  {\path{doi:http://dx.doi.org/10.1016/j.pneurobio.2011.09.005}}.

\bibitem{bgsp}
A.~J. Holmes, M.~O. Hollinshead, T.~M. O’Keefe, V.~I. Petrov, G.~R. Fariello,
  L.~L. Wald, B.~Fischl, B.~R. Rosen, R.~W. Mair, J.~L. Roffman, J.~W. Smoller,
  R.~L. Buckner, {Brain Genomics Superstruct Project} initial data release with
  structural, functional, and behavioral measures, Scientific Data 2 (2015)
  150031.
\newblock \href {http://dx.doi.org/https://doi.org/10.1038/sdata.2015.31}
  {\path{doi:https://doi.org/10.1038/sdata.2015.31}}.

\bibitem{slim}
W.~Liu, D.~Wei, Q.~Chen, W.~Yang, J.~Meng, G.~Wu, T.~Bi, Q.~Zhang, X.-N. Zuo,
  J.~Qiu, Longitudinal test-retest neuroimaging data from healthy young adults
  in southwest {China}, Scientific Data 4 (2017) 170017.
\newblock \href {http://dx.doi.org/https://doi.org/10.1038/sdata.2017.17}
  {\path{doi:https://doi.org/10.1038/sdata.2017.17}}.

\bibitem{maclaren}
J.~Maclaren, Z.~Han, S.~B. Vos, N.~Fischbein, R.~Bammer, Reliability of brain
  volume measurements: A test-retest dataset, Scientific Data 1.
\newblock \href {http://dx.doi.org/https://doi.org/10.1038/sdata.2014.37}
  {\path{doi:https://doi.org/10.1038/sdata.2014.37}}.

\bibitem{sled1998n3}
J.~G. Sled, A.~P. Zijdenbos, A.~C. Evans, A nonparametric method for automatic
  correction of intensity nonuniformity in {MRI} data, IEEE Transactions on
  Medical Imaging 17~(1) (1998) 87--97.
\newblock \href {http://dx.doi.org/https://doi.org/10.1109/42.668698}
  {\path{doi:https://doi.org/10.1109/42.668698}}.

\bibitem{segonne2004skull}
F.~Ségonne, A.~M. Dale, E.~Busa, M.~Glessner, D.~Salat, H.~K. Hahn, B.~Fischl,
  A hybrid approach to the skull stripping problem in {MRI}, NeuroImage 22~(3)
  (2004) 1060--1075.
\newblock \href
  {http://dx.doi.org/https://doi.org/10.1016/j.neuroimage.2004.03.032}
  {\path{doi:https://doi.org/10.1016/j.neuroimage.2004.03.032}}.

\bibitem{jenkinson2002flirt}
M.~Jenkinson, P.~Bannister, M.~Brady, S.~Smith, Improved optimization for the
  robust and accurate linear registration and motion correction of brain
  images, NeuroImage 17~(2) (2002) 825--841.
\newblock \href {http://dx.doi.org/https://doi.org/10.1006/nimg.2002.1132}
  {\path{doi:https://doi.org/10.1006/nimg.2002.1132}}.

\bibitem{addl}
X.~Feng, J.~Yang, Z.~C. Lipton, S.~A. Small, F.~A. Provenzano, Deep learning on
  {MRI} affirms the prominence of the hippocampal formation in {Alzheimer's}
  disease classification, bioRxiv (2018) 456277\href
  {http://dx.doi.org/10.1101/456277} {\path{doi:10.1101/456277}}.

\bibitem{vgg}
K.~Simonyan, A.~Zisserman, Very deep convolutional networks for large-scale
  image recognition, in: International Conference on Machine Learning (ICLR),
  2015.

\bibitem{byrd2019effect}
J.~Byrd, Z.~Lipton, What is the effect of importance weighting in deep
  learning?, in: International Conference on Machine Learning, 2019, pp.
  872--881.

\bibitem{bfrt}
A.~L. Benton, A.~B. Sivan, K.~d. Hamsher, N.~R. Varney, O.~Spreen,
  Contributions to neuropsychological assessment: A clinical manual, Oxford
  University Press, USA, 1994.

\bibitem{fs_thickness}
B.~Fischl, A.~M. Dale, Measuring the thickness of the human cerebral cortex
  from magnetic resonance images, Proceedings of the National Academy of
  Sciences 97~(20) (2000) 11050--11055.

\bibitem{cam}
B.~Zhou, A.~Khosla, A.~Lapedriza, A.~Oliva, A.~Torralba, Learning deep features
  for discriminative localization, in: IEEE Conference on Computer Vision and
  Pattern Recognition (CVPR), 2016, pp. 2921--2929.
\newblock \href
  {http://dx.doi.org/http://doi.computer.org/10.1109/CVPR.2016.319}
  {\path{doi:http://doi.computer.org/10.1109/CVPR.2016.319}}.

\bibitem{gradcam}
R.~R. Selvaraju, M.~Cogswell, A.~Das, R.~Vedantam, D.~Parikh, D.~Batra,
  {Grad-CAM}: Visual explanations from deep networks via gradient-based
  localization, in: IEEE International Conference on Computer Vision (ICCV),
  2017, pp. 618--626.
\newblock \href
  {http://dx.doi.org/http://doi.computer.org/10.1109/ICCV.2017.74}
  {\path{doi:http://doi.computer.org/10.1109/ICCV.2017.74}}.

\bibitem{nam}
X.~Feng, J.~Yang, A.~F. Laine, E.~D. Angelini, Discriminative localization in
  {CNNs} for weakly-supervised segmentation of pulmonary nodules, in:
  International Conference on Medical Image Computing and Computer-Assisted
  Intervention (MICCAI), Springer International Publishing, 2017, pp. 568--576.
\newblock \href
  {http://dx.doi.org/https://doi.org/10.1007/978-3-319-66179-7_65}
  {\path{doi:https://doi.org/10.1007/978-3-319-66179-7_65}}.

\bibitem{nn_frontal_aging}
A.~Gazzaley, J.~W. Cooney, J.~Rissman, M.~D'Esposito, Top-down suppression
  deficit underlies working memory impairment in normal aging, Nature
  Neuroscience 8 (2005) 1298.
\newblock \href {http://dx.doi.org/10.1038/nn1543} {\path{doi:10.1038/nn1543}}.

\bibitem{fdg_frontal_aging_chetelat}
G.~Chetelat, B.~Landeau, E.~Salmon, I.~Yakushev, M.~A. Bahri, F.~Mezenge,
  A.~Perrotin, C.~Bastin, A.~Manrique, A.~Scheurich, M.~Scheckenberger,
  B.~Desgranges, F.~Eustache, A.~Fellgiebel, Relationships between brain
  metabolism decrease in normal aging and changes in structural and functional
  connectivity, NeuroImage 76~(1) (2013) 167--177.
\newblock \href {http://dx.doi.org/10.1016/j.neuroimage.2013.03.009}
  {\path{doi:10.1016/j.neuroimage.2013.03.009}}.

\bibitem{fdg_frontal_aging}
S.~P. Shamchi, M.~Khosravi, R.~Taghvaei, S.~Emamzadehfard, K.~Paydary,
  W.~Raynor, M.~Z. Zadeh, S.~Castro, A.~Nielsen, T.~Werner, Alteration of
  normal regional brain fdg uptake in normal aging, Journal of Nuclear Medicine
  58~(supplement 1) (2017) 483--483.

\end{thebibliography}

\clearpage

\section*{Supplementary Material}
\setcounter{table}{0}
\setcounter{figure}{0}
\renewcommand{\thetable}{S\arabic{table}}
\renewcommand\thefigure{S\arabic{figure}}

\begin{table}[!hbtp]
	\centering
	\caption{\small{Distribution of number of scans per subject}}
	\label{Table:numscanpersubj}
	\begin{tabular}{l|l|l|l|l|l|l|l}
	\hline
	number of scans per subject  & 1       & 2  & 3  & 4  & 5  & 6   &7\\ \hline
	number of subjects          & 2598    & 59 & 19 & 6  & 6  & 3  & 1\\
	\hline
	\end{tabular}
\end{table}

\begin{sidewaystable}
	\centering
		\caption{\small{Dataset information}}
		\label{Table:dataset_name}
		\scriptsize{
			\renewcommand{\arraystretch}{1.1}
		\begin{tabular}{l|l|l}
			Dataset	& Full project name & Source \\ \hline
			ADNI		& Alzheimer's Disease Neuroimaging Initiative  &  \url{http://adni.loni.usc.edu}\\ \hline 
			AIBL			& Australian Imaging Biomarkers and Lifestyle Study of Ageing  &  \url{https://aibl.csiro.au} \\ \hline
			NIFD 		& Frontotemporal Lobar Degeneration Neuroimaging Initiative  &  \url{https://ida.loni.usc.edu/login.jsp?project=NIFD} \\ \hline
			IXI			& Information eXtraction from Images  &   \url{https://brain-development.org/ixi-dataset} \\ \hline
			BGSP		& Brain Genomics Superstruct Project &   \url{https://dataverse.harvard.edu/dataverse/GSP}  \\ \hline
			Cam-CAN		& Cambridge Centre for Ageing and Neuroscience  &   \url{http://www.cam-can.org}   \\ \hline
			OASIS-1		& Open Access Series of Imaging Studies-1  &   \url{https://www.oasis-brains.org} \\ \hline
			OASIS-2		& Open Access Series of Imaging Studies-2  &   \url{https://www.oasis-brains.org} \\ \hline
			SALD		& Southwest University Adult life-span Dataset  &   \url{http://fcon_1000.projects.nitrc.org/indi/retro/sald.html} \\ \hline
			SLIM			& Southwest University Longitudinal Imaging Multimodal Brain Data Repository  &    \url{http://fcon_1000.projects.nitrc.org/indi/retro/southwestuni_qiu_index.html} \\ \hline
			PPMI		& Parkinson's Progression Markers Initiative &   \url{https://www.ppmi-info.org/}  \\ \hline
			SchizConnect 	& SchizConnect  &   \url{http://schizconnect.org/} \\ \hline
			DLBS		& Dallas life-span Brain Study  &   \url{http://fcon_1000.projects.nitrc.org/indi/retro/dlbs.html}  \\ \hline
			CoRR		& Consortium for Reliability and Reproducibility  &   \url{http://fcon_1000.projects.nitrc.org/indi/CoRR/html/}  \\ \hline
	\end{tabular}}
\end{sidewaystable}

\end{document}